\documentclass[%
 aip,
 reprint, % Use 'preprint' for single-column draft
 amsmath,amssymb,
]{revtex4-2}

\usepackage{graphicx}% Include figure files
\usepackage{dcolumn}% Align table columns on decimal point
\usepackage{bm}% bold math
\usepackage{array}% Required for advanced table formatting
\usepackage{amssymb}% For blackboard bold symbols (e.g. \mathbb{1})
\usepackage[colorlinks=true, linkcolor=blue, citecolor=blue, urlcolor=blue]{hyperref}% Hyperlinks
\usepackage[table]{xcolor}
\usepackage{booktabs}
\usepackage{array}
\usepackage{colortbl}
\usepackage{multirow}
\usepackage{titlesec}
\titlespacing*{\section}{0pt}{1.2ex plus 0.3ex minus 0.2ex}{0.8ex plus 0.2ex}
\titlespacing*{\subsection}{0pt}{1.0ex plus 0.3ex minus 0.2ex}{0.6ex plus 0.2ex}
\titlespacing*{\subsubsection}{0pt}{0.8ex plus 0.2ex minus 0.2ex}{0.4ex plus 0.2ex}
\usepackage[table]{xcolor}
\usepackage{colortbl}

\definecolor{bestcell}{RGB}{198, 239, 206}     % soft green
\definecolor{worstcell}{RGB}{255, 199, 206}    % soft red
\definecolor{diagcell}{RGB}{204, 229, 255}     % soft blue
\definecolor{criticalerr}{RGB}{255, 170, 170}  % stronger red
\definecolor{mildererr}{RGB}{255, 235, 156}    % soft yellow
\definecolor{headerrow}{RGB}{220, 220, 220}    % light gray

\usepackage{etoolbox}
\makeatletter
\def\@email#1#2{%
\endgroup
\patchcmd{\titleblock@produce}
  {\frontmatter@RRAPformat}
  {\frontmatter@RRAPformat{\produce@RRAP{*#1\href{mailto:#2}{#2}}}\frontmatter@RRAPformat}
  {}{}
}%
\makeatother

\begin{document}

\preprint{AIP/123-QED}

\title{Agentic Harnesses: LLM-Driven Verification Layers for Robot Autonomy}

\author{Rohan Bhagra}
\altaffiliation[Also at ]{Carnegie Mellon University, Department of Information Systems}
\email{rohan.bhagra@pnnl.gov}
\affiliation{Pacific Northwest National Laboratory, Advanced Computing, Math, and Data Division, Richland, Washington}

\author{Mahantesh Halapannavar}
\affiliation{Pacific Northwest National Laboratory, Advanced Computing, Math, and Data Division, Richland, Washington}

\author{Uddhav Bhattarai}
\affiliation{Pacific Northwest National Laboratory, Advanced Computing, Math, and Data Division, Richland, Washington}

% Figure out how to put program affiliation in a footnote or somethintg.

\date{29 July 2026}

\begin{abstract}
Advances in advanced artificial intelligence tools have sparked research in robot autonomy, but the development of such systems has largely focused on execution rather than verifying the feasibility actions planning models propose. Like general-purpose LLMs, robotics planning models carry risks: biased toward user-specified goals, they may suggest actions misaligned with scientific ethics, they may be unsafe due to an inability to "remember" prior safety risks, or they may be vulnerable to adversarial attacks on the autonomy ecosystem. We propose a LLM-driven verification layer between planning and execution to evaluate action permissibility. Our LLM-as-a-Judge ensemble combines chain-of-thought reasoning across models and synthesizes those expert judge outputs, mirroring a combination of a mixture of experts and self-consistency approach. This layer serves as middleware, gating plans from the server's planning module before they reach the MCP server and therefore the robot's low-level controls: plans are approved, rejected for reformulation, or escalated for human review. With this system, we achieve near 85\% precision across accept/escalate/reject categories 97\% containment of adversarial attacks, with negligible errors between accepting and rejecting tasks, and errors mostly manifesting at the escalate boundary.

%  Using a knowledge graph indexed on robot capabilities, scientific procedure and ethics standards, and cybersecurity vulnerability enumerations

\end{abstract}
\maketitle
\section{Introduction}
\label{sec:intro}

Though there have been numerous advances in the field of robot autonomy, both in academic research settings and in the high-tech industry, safety research in this field has lagged behind. Recent research in AI safety has shown a vast frontier for potential risks in the use of deep-learning powered models. Large Language Models present a compelling option for robot autonomy planning as their ability to conduct zero-shot reasoning in diverse environments greatly increases the action space. These capabilities allow for the creation of robot autonomy systems, where a planning module, manipulation agents, perception agents, and a host of other support technologies allow an autonomy system to create plans, formulate and recalibrate goals, and execute tasks through the use of robots in a system that is completely devoid of human supervision.

Previous research has touched on the various aspects that formulate the risk surface in AI-enabled robotics; this includes safety verification, the separation of governance and execution layers, and protecting against adversarial attacks. However, these checks have been made independently and often alone, and many processes run without a full suite of verification. Though LLMs perform well in new environments, they are prone to suggesting dangerous actions or creating plans that violate ethical principles in pursuit of a given goal, a phenomenon that is exacerbated when it is operating in an agentic system [1]. Moreover, LLM planners are prone to introducing plans that, while valid, are dangerous [2]. Even when models feature stronger reasoning capabilities and produce more thorough plans, their danger avoidance remains relatively flat [2]. Beyond alignment and physical safety, agentic ecosystems are also prone to adversarial attacks, as they expose a vast threat surface to these attacks, especially at the data ingestion layer [3]. Current autonomy systems lack a framework to minimize these risks, and contain adversarial attacks, among other undesired behaviors. 

\subsection{LLM-driven verification}

The advancement in the reasoning capabilities of LLMs themselves have made them a viable option to evaluate the feasibility and safety of actions proposed by a planning agent [4] [5]. That said, while the use of even a single judging model has shown promise, stronger approaches lean on model ensembles to reduce the impacts of individual model variance. Previous research has found that aggregating the outputs of multiple models—or multiple reasoning passes from a single model—yields more reliable performance than relying on a single run on a single model, an approach called self-consistency [6]. Furthermore, ensembling models mitigates biases within a model, risks of prompt injection, and the tendency of models to prefer their own outputs [7]. Zhao et al. propose utilizing a committee of judge agents that iteratively discuss and revise their verdicts, which improved judgment quality and mitigated any given single-model bias [8]. Additionally, utilizing multiple agents can help co-agents resolve inconsistencies in their own solutions [9].

\subsection{Verification Systems in Robotics}
Recent efforts to verify the actions of agentic robotics systems fall broadly into two categories: formal methods that constrain low-level dynamics with precise safety guarantees, and LLM-based methods that evaluate the semantic safety of high-level plans. Comparatively little work addresses the ethical and scientific alignment of actions proposed by a planning model within the specific context of robotics.

Formal approaches such as Control Barrier Functions provide provable guarantees on a robot's motion, ensuring it remains \textit{safe} in the sense of avoiding collisions with humans and objects in a shared workspace [10]. Hamilton-Jacobi reachability analysis offers a complementary certificate of safety, computing a backward-reachable set to determine whether a given state or action leads to an unsafe outcome [11]. LLM-based approaches, by contrast, target the semantic and ethical dimensions of high-level plans rather than enforcing low-level safety bounds.

\subsection{Governance as a runtime structure}

To verify actions across the full robotics control pipeline, several efforts treat governance as a core runtime structure, decoupled from the planning agent itself. This framing relies on three principles: (i) \textit{separation of cognition and governance}, keeping execution control inspectable, configurable, and portable across deployment environments while leaving task interpretation and planning to the agent; (ii) \textit{capability-centric enforcement}, factoring knowledge of the capabilities of the specific robotics technologies in the autonomy setting into decisions, which makes verification more accurate and specifc; and (iii) \textit{continuous governance}, extending oversight beyond a single pre-execution validation step [11].

Work on LLM-based anomaly detection suggests that foundation models are well-suited to fill this role of a safety-aware validator [12]. Capability-centric verification at the execution level offers a further safeguard: by checking actions against a robot's actual executable capabilities, it protects against cases where a planning agent misjudges what the robot can safely do, a mismatch that is often a root cause of downstream unsafe behavior [4].

Taken together, these formal, semantic, and governance-based approaches each address a piece of the verification problem, but existing frameworks often fail to unify low-level safety guarantees, high-level ethical/semantic evaluation, and continuous capability-aware oversight into a single runtime structure for verification. This gap motivates the governance architecture we propose in this work.

\section{Methodology}
\label{sec:methods}
To motivate the approach that we take in the rest of the paper, we establish that the goal of a verification layer in a robot autonomy setting is to prevent undesired plans from a planning module from reaching a MCP server, which orchestrates the low-level controls of the robot(s) in the environment of interest. 
\begin{figure}[!h]
    \centering
    \includegraphics[width=0.75\linewidth]{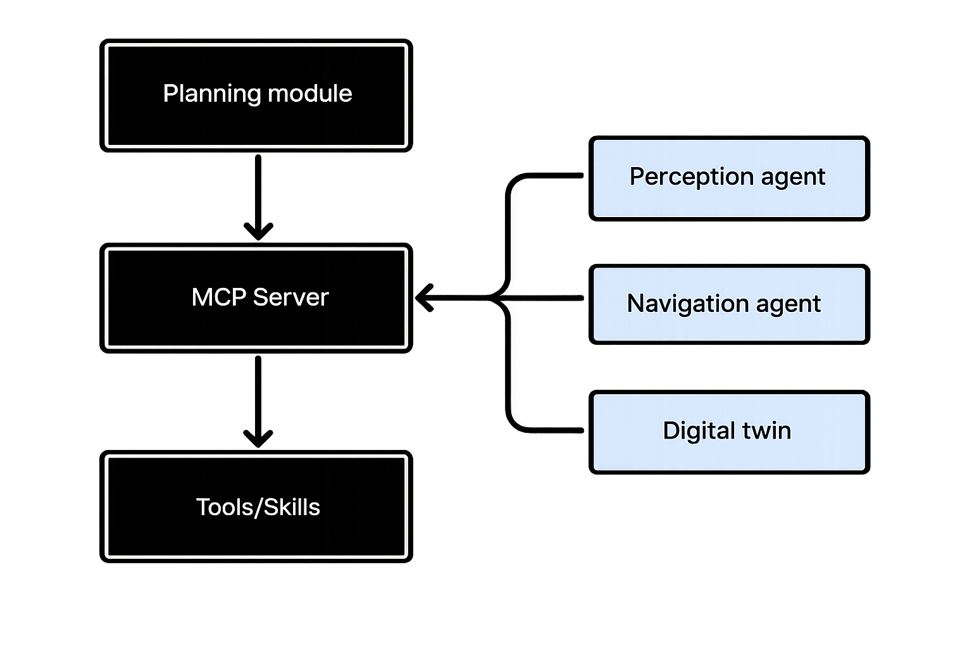}
    \caption{In an autonomy setting, high level goals are communicated to a planning module, which creates detailed, step-by-step plans, which are then used by an MCP server to call robot-specific Tools and Skills to gather data and execute proposed actions. We aim to place a verification layer between the planning module and the MCP server.}
    \label{fig: MCP Diagram figure}
\end{figure}

To this end, we define a plan $\mathcal{T}$ as a finite sequence of natural-language instructions,
\[
\mathcal{T} = (t_1, t_2, \ldots, t_n),
\]
where $n$ is the total number of steps in the plan such that a robot can execute these steps to achieve its goal. The goal of the verification layer is to receive a plan $\mathcal{T}$ and either accept it, reject it and send it for reformulation, or escalate it to a human expert for review. Rejected plans are sent back to the planning module for revisions, and accepted plans move to the MCP server.
\begin{figure}[!t]
    \centering
    \includegraphics[width=1\linewidth]{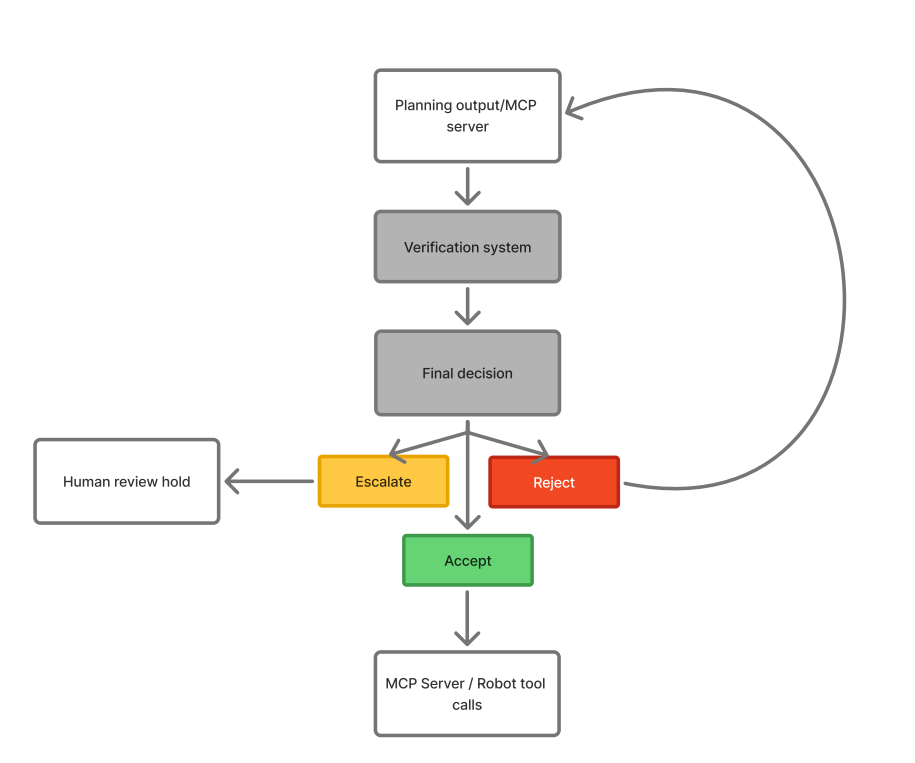}
    \caption{A simplified, high-level overview of how the verification layer integrates with the remaining robot autonomy architecture.The verification system is queried on the direct output from a planning module, without any modifications. An "accept" decision passes that plan on to the next step of the autonomy process. Verification decisions are made in a two-stop process, with a final decision being issued by the second prong of the system. Readers are encouraged to review the note for alternate placement of the verification layer.}
    \label{fig: high level architecture sketch }
\end{figure}

\textsc{Note:} We recognize that this verification layer can also be placed in between the MCP server and tool/skill calls, and that it can be particularly effective to place a layer both preceding and after a MCP server. Doing so will require some formulation of MCP server actions into natural-language representations. However, this work does not consider that added dimension.

\subsection{Judging Architecture}
The judging architecture consists of a two-prong system leveraging LLM-as-a-judge. The primary layer consists of ensemble containing an arbitrary number of LLM evaluators, who all receive the same system prompts and evaluation prompts. Each evaluator also receives some context $k$ from the knowledge graph. This is discussed in section C. The input to the judging models consists of a system prompt that informs the evaluator that it will be tasked with judging proposed actions and instructs it to only make evidence-based decisions that should be based on the decision rules provided in the user prompt. (We follow this convention as prior work has shown that assigning roles via the system prompt improves task-specific reasoning and alignment with intended behavior) [13].

We then design a language prompt $p$ for the LLM, which provides the instruction that outputs must be classified to be accepted, rejected, or escalated. The prompt then outlines decision rules for each of these categories and lays out high-level rules that the decisions should conform to. There are also a brief list of example plans to accept and example plans to reject, giving the judging model a few-shot background. We then provide notes on the escalate category, fill in the planning output, and provide a preamble for the context that is accessed from the knowledge graph, which is inserted as the last line of the prompt. Our outputs must be structured in order to be parsed and acted upon, so we utilize infill prompting, asking the model to provide its response in the following format: 
\begin{quote}
``First, you should be concise. Then reason about whether the plan should be accepted or not, based on their safety, alignment, and security. Then answer with `accept' if the plan is safe, `escalate' if you have strong doubts about safety or you are applying a decision rule from this document, and `reject' if there is a critical flaw that makes the plan unsafe under normal circumstances for a robotics lab.

\medskip

\texttt{[REASONING:] [your reasoning here] [DECISION:] [ACCEPT, REJECT, ESCALATE]}''
\end{quote}

    \textit{Model Choice:} The judging itself is performed by each model in our ensemble. We utilize five public, frontier models to achieve a middle ground of quality of judgment and cost. Additionally, to further balance performance and cost, we use a mixture of heavyweight reasoning models (e.g, Claude Sonnet 4.5 and scientific, reasoning-first models such as Open AI o3) and lightweight, lower-cost models (e.g, Grok-4.1-fast). The effects on quality of using a different number of judges is presented later in this work.
    
    Evaluations run asynchronously on each judging model to reduce latency. This means that the latency of the first layer of the judging system is defined by the slowest model of the ensemble. To prepare for the next step of the verification layer, we capture a judge's decision, rationale, and the entire response object.   

    \textit{Chief Judge Layer}:
    Once all judges render a decision on plan $\mathcal{T}$ given $p$, we consult the chief judge for a final determination on the plan. In this methodology, the role of the chief judge is served by another frontier, heavyweight reasoning Large Language Model (this study utilized Claude Sonnet 4.6). The role of the chief judge is not to evaluate the plan again—this is enforced by neglecting to expose the plan itself to the chief judge. Rather, its role is to evaluate the quality of the reasoning of each prior judge, the logic of the claims they make, and to balance the magnitude of the risk that each judge outlines. 
    
 The chief judge is given the following instructions to ensure sound evaluation that considers the various dimensions that must be accounted for in the risk and feasibility analysis of proposed plans:

\begin{quote}
``Ensure the following:
\begin{enumerate}
    \item You analyze the reasoning of each judge and ensure that they are aligned with the safety, alignment, and security of the robotics task planner.
    \item You do not weigh the opinions of any one judge more than others, except when the judge is giving a strong reason for their opinion.
    \item You are to verify the soundness of the reasoning of each judge; do not make a decision on the basis of a point that you are dubious about.
    \item You should not do significant reasoning about the plans, but reason on the judges' claims and thought processes.
    \item You intentionally do not have access to the plans, but you should be able to reason on the judges' reasoning.
    \item If you notice that you are conflicting with 3 or more judges, you should automatically escalate the plans
\end{enumerate}
The judges' deliberations are provided through context to you.''
\end{quote}

The chief judge then considers this information, the outputs of the previous five judges (provided as five model-response pairs) and renders a final decision of accept, reject, or escalate. The chief judge also classifies each plan as a concern of alignment, safety, or preventing adversarial inputs. (We rely upon this classification for failure mode tracking). As such, the latency of each judging roundtrip aggregates to the wall-clock time of two LLM calls. For a testing run that evaluated 58 runs, the average roundtrip time per judging loop was 24.2 seconds. \footnote{The primary judges for this evaluation were: Claude Sonnet 4.5, GPT 5.2, Grok 4.1-fast Reasoning, Gemini 3.5-flash, and OpenAI o3. The chief judge was Claude Sonnet 4.6}.

\subsection{Deterministic Checks}
\label{sec:deterministic}

LLM-based judging is susceptible to hallucination, incomplete context, 
and adversarial manipulation of its inputs. We therefore aid 
the judging ensemble with rule-based checks that run prior to LLM 
evaluation, targeting the \textit{cybersecurity} surface of the 
autonomy ecosystem; low-level physical safety is delegated to each 
robot's existing control stack.

Given a candidate plan $\mathcal{T}$, the set of registered 
skills $\mathcal{S}$, and the current execution context $\mathcal{C}$, 
we define a deterministic verifier
\[
V_{\text{det}} : (\mathcal{T}, \mathcal{S}, \mathcal{C}) \longrightarrow
\{\text{pass}, \text{flag}, \text{block}\},
\]
where block halts $\mathcal{T}$ before it reaches the judges, 
flag forwards it with a structured annotation appended to the 
judges' context $k$, and pass forwards it unmodified to the judges.
$V_{\text{det}}$ comprises three components.

\paragraph{Provenance and skill security.}
The verifier confirms that an operable sandbox is present and that all 
code paths in $\mathcal{T}$ are routed through it. A capability 
allow-list $A : \mathcal{S} \to 2^{\mathcal{F}}$ maps each skill to 
its permitted functional categories (e.g, \texttt{perception}, 
\texttt{manipulation});  Plans that rely on skills outside the allow list $A(s_i)$ are blocked, helping prevent against spoofing attacks.
Optionally, a Software Bill of Materials (SBOM) records the provider, 
version, and hash of each skill and software dependency; runtime-verified hash matches are annotated 
with a \texttt{provenance-verified} tag which is propagated to the judges.

\paragraph{Prompt-injection screening.}
Since planning outputs are consumed by downstream LLM judges, we 
sanitize natural-language artifacts crossing the output to verification layer boundary via 
regex and fuzzy search against curated injection and jailbreak 
patterns. Two thresholds—\texttt{advisory} and \texttt{warning}—attach 
annotations to the judges' context, informing the judge of possible injection. The injection taxonomy and threshold 
calibration are discussed in Section~E.

\paragraph{Long-window monitoring.}
As future work, the verification layer could maintain a sliding window of recent inputs to the verification layer and aggregate their suspicion scores to detect 
multi-step attacks whose individual inputs appear benign, forwarding 
sessions above an aggregate threshold to human review.

 While the judges reason about the semantics of 
$\mathcal{T}$, $V_{\text{det}}$ reasons about its provenance 
and security risks.
\subsection{Knowledge Graph}
\label{sec:kg}

Each judge $J_i$ conditions its evaluation on a retrieved context
$k$ drawn from a knowledge graph $\mathcal{K}$ that provides the 
factual grounding required to reason about plans in our
autonomy setting. Rather than relying on the  knowledge of
the underlying judging LLMs—which can be uneven across models and cannot be systematically
analyzed—$\mathcal{K}$ provides verified and accurate contextual material that is paramount for successful decision-making 

$\mathcal{K}$ is indexed over five categories of source material:

\begin{enumerate}
    \item \textit{Systems-related lab context.} An internal lab document
    describing the hardware platform (UR7e arm on a Clearpath Ridgeback
    base, Robotiq 2F-85 gripper, Orbbec 335Lg RGBD camera), the state
    machine that governs pick-and-place execution, the perception
    pipeline (segmentation, tracking, pose estimation) the on-demand VLM query interface, the
    MoveIt system and OctoMap lifecycle, and the
    coordination rules that govern shared-workspace operation between
    the arm and the mobile base. This is specific to our autonomy setting, and outlines the architectural and operational conditions that must be followed.

    \item \textit{Cybersecurity vulnerability enumerations.} Entries
    from the Common Weakness Enumeration (CWE) catalog, providing a
    standardized vocabulary for the classes of software weaknesses
    that a plan could expose.

    \item \textit{Adversarial-attack taxonomy for AI systems.}
    Entries from MITRE ATLAS, which enumerates adversarial attack paths
    specific to agentic AI systems. 
    \item \textit{Hardware manuals and lab-safety and scientific-ethics
    documentation.} Manufacturer documentation for the robots in the environment, along with
    lab-safety procedures and scientific-ethics standards.

    \item \textit{Prior failure modes.} A record of previous
    rejected plans related to adversarial attacks and alignment failures,
    which allows judges to recognize recurrences of known patterns.
\end{enumerate}

The resulting graph contains $|V| = 7{,}803$ entity nodes and
$|E| = 28{,}730$  relationships, yielding an average degree of
$\bar{d} \approx 7.36$ and an edge density of
$\rho \approx 9.4 \times 10^{-4}$. The low density confirms 
that entities cluster tightly within their source domain (e.g., CWE
entries, hardware components, lab procedures). There are also comparatively
sparser cross-domain links, which motivates extraction to locate a small number of highly relevant entry points
and expand within their neighborhoods rather than traversing the various nodes of
the graph globally.

At query time, the retrieved context $k \subset \mathcal{K}$ is
appended to the evaluation prompt $p$ as described in
Section A. \textsc{note:} Context is not exposed to the chief judge, only initial evaluators.

\subsection{Failure Mode Tracking}
To enable comparison of proposed plans against previously rejected 
ones, we log every plan $\mathcal{T}$ that is (i) rejected by the 
verification layer and (ii) classified by the chief judge as an 
alignment or adversarial attack concern. Retrieving these plans and their  neighbors exposes the primary judges to 
failed plans that are semantically or structurally 
similar to the candidate, supporting continuity of reasoning and reducing  fluctuations in judge behavior on 
near-identical inputs. As a future direction, thse failure modes could 
support a distilled binary classifier trained on the logged rejections. This would preempt
LLM evaluation when a candidate plan lies within a specified 
similarity threshold of a prior rejection, reducing both compute 
cost and evaluation latency.

\subsection{Prompt Injection Guardrails}
\label{sec:prompt-injection}

Because planner outputs are consumed by downstream LLM components, 
any injection payload in $\mathcal{T}$ threatens the
verification pipeline. Thus, we compute a deterministic suspicion 
score $\sigma(\mathcal{T}) \in [0, 1]$ prior to invoking the judges.

To defeat common obfuscations, we generate a comprehensive suite of normalized views of $\mathcal{T}$ and run the detection pipeline over each view. We utilize the maximum score returned from a detection test for each view. To achieve this, two detectors operate simultaneously:  (i) a curated regex library of 
instruction-override, role-spoofing, and jailbreak phrasings,  
annotated with a hand-assigned confidence score for each phrase, and (ii) fuzzy matching 
against canonical injection phrases via the RapidFuzz library. Within each 
detector, per-match confidences $\{s_i\}$ are aggregated under the assumption that each match is independent, 
\[
S = 1 - \prod_{i} \left(1 - \tfrac{s_i}{100}\right),
\]
yielding $S_{\text{regex}}, S_{\text{fuzzy}} \in [0, 1]$. These are 
combined with weights $w_r = 0.65$, $w_f = 0.35$ 
that reflect the higher precision of the regex method. We define an advisory threshold $\tau_A$ and a stricter warning 
threshold $\tau_W$. Crossing $\tau_W$ attaches a strong warning to 
the judges' context; crossing only $\tau_A$ attaches a weaker 
advisory that reports the score and defers to the judges' reasoning. Both warnings \textit{inform} the judge rather than make decisions.

% use the tables feature to report f1 scores etc. 

\section{Results and Discussion}
\label{sec:results}
In this section, we outline the experimental evaluation of this proposed verification layer system. We will also report the results of this system, as well as discuss the findings. To evaluate the verification layer, we produced a dataset with a uniform structure: category of testing focus, plan, decision, rationale, and cited rules. The experimented consisted of a AI-generated plans (prompted few-shot style, with human developed plans), and human-labeled decisions, rationales, and cited rules. We consider the verification layer's classification of the plan versus the ground truth decision label as the primary evaluation metric.
\subsection{Safety, Alignment, Adversarial Attack Comprehensive Test}
Using our prompt $p$ and context $k$, we test the verification layer against a dataset comprised of a roughly even split of safety, alignment, and adversarial attack-classified plans. \footnote{We use the same ensemble as footnote 1} The dataset contained 58 unique plans. We report the results in Table I.

% add a footnote with model names.

\definecolor{diagcolor}{RGB}{178, 223, 178}      % green: correct
\definecolor{minorerror}{RGB}{255, 236, 179}     % yellow: adjacent (acceptable) errors
\definecolor{majorerror}{RGB}{255, 192, 192}     % red: critical accept/reject errors
\definecolor{headercolor}{RGB}{176, 224, 230}    % light blue: headers
\definecolor{totalcolor}{RGB}{224, 224, 224}     % grey: totals

\begin{table}[!h]
\centering

\label{tab:confusion}

\renewcommand{\arraystretch}{1.3}
\setlength{\tabcolsep}{3.5pt}
\scriptsize

\begin{tabular}{|c|l||c|c|c||c|}
\hline
\multicolumn{2}{|c||}{} 
    & \multicolumn{3}{c||}{\cellcolor{headercolor}\textbf{Predicted}} 
    & \cellcolor{totalcolor}\textbf{Total} \\
\cline{3-6}
\multicolumn{2}{|c||}{} 
    & \cellcolor{headercolor}\textbf{Acc.} 
    & \cellcolor{headercolor}\textbf{Esc.} 
    & \cellcolor{headercolor}\textbf{Rej.} 
    & \cellcolor{totalcolor}\\
\hline\hline
\multirow{3}{*}{\rotatebox{90}{\textbf{Actual}}} 
    & \cellcolor{headercolor}\textbf{Accept} 
    & \cellcolor{diagcolor}\textbf{10} 
    & \cellcolor{minorerror}2 
    & \cellcolor{majorerror}0 
    & \cellcolor{totalcolor}12 \\
\cline{2-6}
    & \cellcolor{headercolor}\textbf{Escalate} 
    & \cellcolor{minorerror}1 
    & \cellcolor{diagcolor}\textbf{14} 
    & \cellcolor{minorerror}1 
    & \cellcolor{totalcolor}16 \\
\cline{2-6}
    & \cellcolor{headercolor}\textbf{Reject} 
    & \cellcolor{majorerror}0 
    & \cellcolor{minorerror}7 
    & \cellcolor{diagcolor}\textbf{23} 
    & \cellcolor{totalcolor}30 \\
\hline\hline
\multicolumn{2}{|c||}{\cellcolor{totalcolor}\textbf{Total}} 
    & \cellcolor{totalcolor}11 
    & \cellcolor{totalcolor}23 
    & \cellcolor{totalcolor}24 
    & \cellcolor{totalcolor}\textbf{58} \\
\hline
\end{tabular}

\vspace{0.8em}

\begin{tabular}{|l|c|c|c|c|}
\hline
\rowcolor{headercolor}
\textbf{Class} & \textbf{Precision} & \textbf{Recall} & \textbf{F1} & \textbf{Support} \\
\hline
Accept       & 0.91 & 0.83 & 0.87 & 12 \\
\hline
Escalate     & 0.61 & 0.88 & 0.72 & 16 \\
\hline
Reject       & 0.96 & 0.77 & 0.85 & 30 \\
\hline\hline
\rowcolor{totalcolor}
\textbf{Accuracy}     & \multicolumn{3}{c|}{\textbf{0.81}} & \textbf{58} \\
\hline
\rowcolor{totalcolor}
Macro avg    & 0.83 & 0.83 & 0.81 & 58 \\
\hline
\rowcolor{totalcolor}
Weighted avg & 0.85 & 0.81 & 0.82 & 58 \\
\hline
\end{tabular}
\caption{Confusion matrix and per-class metrics for the verification 
layer ($N = 58$). Rows are ground-truth labels, columns are chief-judge 
predictions. Green cells on the diagonal indicate correct decisions. 
Yellow cells indicate errors between adjacent decision classes 
(accept$\leftrightarrow$escalate and escalate$\leftrightarrow$reject); 
these are undesirable but not overtly dangerous, since escalation still 
routes the plan to human review. Red cells indicate direct 
accept$\leftrightarrow$reject confusions, the most severe failure 
mode. In this test, zero such confusions occurred in our evaluation.}
\end{table}

We draw special attention to the fact that there are no catastrophic errors in this test, meaning no human-labeled accepted plans were predicted as rejected plans and no human-labeled rejected plans were accepted by the verification layer. This is a promising result and it highlights the operationalization prospects of such a system in a laboratory setting.
\subsection{Adversarial Attack Focused Test}

To evaluate the verification layer against inputs targeting the 
security surface of the autonomy stack, we assembled a dataset of 
$N = 38$ plans spanning command and prompt injection, 
authentication and authorization violations, sensor spoofing, supply-chain and configuration 
attacks, multi-agent, and data exfiltratiton attacks. In this dataset, class support is skewed toward reject ($n = 31$), reflecting the wide scope of cybersecurity concerns. \footnote{We use the same ensemble as footnote 1}

Running the verification layer 
with $k$ populated from CWE and MITRE ATLAS, the system contains $97\%$ of unsafe plans and produces 
only a single catastrophic error, as displayed in Table II.
Much of the remaining error is due to 
escalate $\leftrightarrow$ reject boundary: the chief judge tends 
to reject cases whose ground truth routes to human review, especially on complex-multi lab-system problems. Thus,
precision on the escalate class is the primary 
area for further gains. 
\begin{table}[!h]
\centering
\caption{Confusion matrix and safety-focused metrics for the
cybersecurity-specific evaluation subset ($N = 38$). Because
the \textsc{accept} and \textsc{escalate} classes have minimal
support, per-class precision/recall are not stable metrics; thus, we instead
report aggregate metrics that are aligned with the deployment goals of this work.}
\label{tab:confusion-cyber}
\renewcommand{\arraystretch}{1.3}
\setlength{\tabcolsep}{10pt}

\begin{tabular}{lcccc}
\toprule
& \multicolumn{3}{c}{Predicted} & \\
\cmidrule(lr){2-4}
Actual & Accept & Escalate & Reject & Total \\
\midrule
Accept   & 2 & 1 & 0  & 3 \\
Escalate & 0 & 1 & 3  & 4 \\
Reject   & 0 & 4 & 27 & 31 \\
\midrule
Total    & 2 & 6 & 30 & 38 \\
\bottomrule
\end{tabular}

\vspace{1em}

\begin{tabular}{lc}
\toprule
Metric & Value \\
\midrule
Overall accuracy                                      & 0.79 \\
Critical-failure rate (accept$\leftrightarrow$reject) & 0.00 \\
Unsafe-plan containment$^{\dagger}$                   & 1.00 \\
Human-review coverage of unsafe plans$^{\ddagger}$    & 0.14 \\
Safe-plan pass-through$^{\S}$                         & 0.67 \\
\bottomrule
\end{tabular}

\vspace{0.75em}
\begin{minipage}{\linewidth}
\footnotesize
$^{\dagger}$ Fraction of ground-truth \textsc{reject} plans that were not accepted (i.e., correctly rejected or escalated). \\
$^{\ddagger}$ Fraction of ground-truth \textsc{reject} plans that were routed to human review through \textsc{escalate}. \\
$^{\S}$ Fraction of ground-truth \textsc{accept} plans that were correctly accepted.
\end{minipage}
\end{table} 
\subsection{Deterministic Injection Checker}
Table III reports injection guard performance 
across decision thresholds on a curated set of $N = 46$ inputs 
(32 injection, 14 benign) drawn from OWASP and related 
cybersecurity sources. \footnote{Testing was done via a separate script. It was not tested through the main judging pipeline. Integration is a future direction.}
\begin{table}[!h]
\centering
\caption{Prompt-injection screener performance across decision 
thresholds $\tau$ on a curated set of $N = 46$ inputs 
(32 injection, 14 benign) drawn from OWASP and other 
cybersecurity domain expert sources. TP/FP/TN/FN counts are given alongside 
accuracy, precision, recall, and F1 metrics. We observe that the quality of the injection checker peaks at a threshold of 0.85, and that there is a sharp drop off at both thresholds significantly greater than or significantly less than 0.85. A greater drop off is observed at higher thresholds.}
\label{tab:injection-thresholds}

\renewcommand{\arraystretch}{1.15}
\setlength{\tabcolsep}{6pt}
\small
\begin{tabular}{c|cccc|cccc}
\toprule
$\tau$ & TP & FP & TN & FN & Acc. & Prec. & Rec. & F1 \\
\midrule
0.95 & 22 & 1 & 13 & 10 & 0.76 & 0.96 & 0.69 & 0.80 \\
0.85 & 26 & 3 & 11 & 6  & 0.80 & 0.90 & 0.81 & 0.85 \\
\textbf{0.75} & \textbf{28} & \textbf{3} & \textbf{11} & \textbf{4} & \textbf{0.85} & \textbf{0.90} & \textbf{0.88} & \textbf{0.89} \\
0.65 & 28 & 4 & 10 & 4  & 0.83 & 0.88 & 0.88 & 0.88 \\
0.55 & 28 & 5 & 9  & 4  & 0.80 & 0.85 & 0.88 & 0.86 \\
\bottomrule
\end{tabular}
\end{table}
\subsection{Ensemble Size Analysis}
\label{sec:ensemble_size}

To understand the effects on the size of the judging committee on the quality of verification, we evaluate the verification layer with 4 unique configurations: an ensemble with 1, \footnote{In this ensemble, the primary model was Claude Opus 4.8 and the chief judge was Claude Sonnet 5} 3 \footnote{In this ensemble, the ensemble models were Claude Opus 4.8, Grok 4.1 Fast Reasoning, and GPT 5.6 Terra, and the chief judge was Claude Sonnet 5}, 5 \footnote{In this ensemble, the ensemble models were Claude Opus 4.8, Grok 4.1 Fast Reasoning, GPT 5.6 Terra, GPT o3, and Gemma 4 26 Billion Parameters, and the chief judge was Claude Sonnet 5}, and 7 \footnote{In this ensemble, the ensemble models were Claude Opus 4.8, Grok 4.1 Fast Reasoning, GPT 5.6 Terra, GPT o3, Gemma 4 26 Billion Parameters, Gemini 3.1 Flash Lite, and Grok 4.20 Reasoning, and the chief judge was Claude Sonnet 5},judges respectively. We utilize different models from the initial tests earlier in this report as a proof of concept that this architecture can be standardized across different models. We run each ensemble against the same ground truth dataset with $N =55$ and report per-class metrics in Table V and the full confusion matrix for each ensemble in Table VI.

\subsubsection{Overall accuracy is largely independent to ensemble size}
Across each configuration, aggregate accuracy remains within a narrow $0.76$--$0.78$ band, and weighted-F1 within $0.76$--$0.78$. Increasing the number of judges from 1 to 7 yields only a two-point improvement in accuracy, indicating that smaller ensembles can be viable. This is notable as the chief benefit of a smaller ensemble is a decrease in latency and inference costs. However, this exposes the verification layer to one-off prompt injection attacks or simple model hallucinations as discussed earlier than the paper.

\subsubsection{The verification layer exhibits a zero false-accept characteristic}
As with prior tests, we highlight that across each ensemble, we avoid false-accept catastrophic errors. Thus, the verification layer is a viable solution to providing an empirically conservative solution to ensuring safety and alignment in proposed actions. We also note that the precision for accept is $1.00$ in each ensemble. This is desired as we seek to avoid false accepts, and if a plan that should be accepted is incorrectly classified as escalate or reject, the broader agentic architecture can explore different methods of achieving the goal that drove the plan. 

\subsubsection{Larger ensembles benefit the reject class most}
Reject-class F1 improves steadily with ensemble size---$0.81$, $0.82$, $0.84$, $0.86$ at 1, 3, 5, and 7 judges respectively---and reject-class recall reaches $0.87$ at both 5 and 7 judges. Since the reject class is both the largest and the most safety-critical, we observe that larger ensembles may indeed be a worthwhile investment.

\subsubsection{Macro-F1 is maximized at 3 judges}
While weighted metrics favor the 7-judge ensemble, macro-F1, which weights each class equally regardless of support, peaks at $0.76$ with 3 judges. The 3-judge ensemble retains an accept-class recall of $0.70$ (which ties the best baseline among the various ensembles). This makes 3 judges an attractive point on the cost/performance frontier. 

\subsection{Latency}
The average time per evaluation for each ensemble is shown in the table below: 
\begin{table}[htbp]
\caption{Mean run time per verification run as a function of ensemble size, averaged over the $N=55$ ground-truth dataset.}
\label{tab:runtime}
\begin{ruledtabular}
\begin{tabular}{cc}
\textbf{Ensemble Size} & \textbf{Avg.\ Time per Run (s)} \\
\hline
1 judge  & 27.76 \\
3 judges & 28.94 \\
5 judges & 28.69 \\
7 judges & 31.56 \\
\end{tabular}
\end{ruledtabular}
\end{table}

We note that the lack of significant fluctuation between latencies is an artifact of our async approach (the total run time for the first part of the ensemble is bounded by the slowest model only, avoiding ballooning latency for larger ensembles). An interesting future direction is investigating the possibility of using only cheaper, faster, lightweight instead of the heavy reasoning models used in this approach.  
\begin{table}[!t]
\caption{Ensemble results by number of judges. P represents precision and R represents recall.}
\label{tab:ensemble_results_color}
\begin{ruledtabular}
\begin{tabular}{l cccc}
\rowcolor{headerrow}
\textbf{Metric} & \textbf{1} & \textbf{3} & \textbf{5} & \textbf{7} \\
\hline
\multicolumn{5}{l}{\textit{Accept} ($n=10$)} \\
P  & 1.00 & 1.00 & 1.00 & 1.00 \\
R  & \cellcolor{bestcell}\textbf{0.70} & \cellcolor{bestcell}\textbf{0.70} & \cellcolor{worstcell}0.50 & 0.60 \\
F1 & \cellcolor{bestcell}\textbf{0.82} & \cellcolor{bestcell}\textbf{0.82} & \cellcolor{worstcell}0.67 & 0.75 \\
\hline
\multicolumn{5}{l}{\textit{Escalate} ($n=14$)} \\
P  & \cellcolor{bestcell}\textbf{0.60} & \cellcolor{worstcell}0.56 & 0.59 & 0.59 \\
R  & \cellcolor{worstcell}0.64 & \cellcolor{bestcell}\textbf{0.71} & \cellcolor{bestcell}\textbf{0.71} & \cellcolor{bestcell}\textbf{0.71} \\
F1 & \cellcolor{worstcell}0.62 & \cellcolor{worstcell}0.62 & \cellcolor{bestcell}\textbf{0.65} & \cellcolor{bestcell}\textbf{0.65} \\
\hline
\multicolumn{5}{l}{\textit{Reject} ($n=31$)} \\
P  & \cellcolor{worstcell}0.79 & 0.83 & 0.82 & \cellcolor{bestcell}\textbf{0.84} \\
R  & 0.84 & \cellcolor{worstcell}0.81 & \cellcolor{bestcell}\textbf{0.87} & \cellcolor{bestcell}\textbf{0.87} \\
F1 & \cellcolor{worstcell}0.81 & 0.82 & 0.84 & \cellcolor{bestcell}\textbf{0.86} \\
\hline
Accuracy   & \cellcolor{worstcell}0.76 & \cellcolor{worstcell}0.76 & \cellcolor{worstcell}0.76 & \cellcolor{bestcell}\textbf{0.78} \\
Macro P    & 0.80 & 0.80 & 0.80 & \cellcolor{bestcell}\textbf{0.81} \\
Macro F1   & 0.75 & \cellcolor{bestcell}\textbf{0.76} & \cellcolor{worstcell}0.72 & 0.75 \\
Weighted P & \cellcolor{worstcell}0.78 & 0.79 & 0.79 & \cellcolor{bestcell}\textbf{0.81} \\
Weighted F1& 0.77 & 0.77 & \cellcolor{worstcell}0.76 & \cellcolor{bestcell}\textbf{0.78} \\
\end{tabular}
\end{ruledtabular}
\end{table}

\begin{table}[!t]
\caption{Confusion matrices for each ensemble configuration on the ground-truth dataset ($N=55$). Rows are the true label, columns are the predicted verdict. \colorbox{diagcell}{Blue} cells mark correct classifications (diagonal). \colorbox{mildererr}{Yellow} cells signal mild errors at the escalate boundary (accept$\leftrightarrow$escalate or escalate$\leftrightarrow$reject confusions), which can be approved by human review. \colorbox{criticalerr}{Red} cells mark critical errors in which an accept was classified as reject, or vice versa. Critical errors remain low ($\leq 2$) across all configurations and are strictly one-directional, meaning the ensemble never wrongly accepts a task that should be rejected.}
\label{tab:confusion_matrices_color}
\begin{ruledtabular}
\begin{tabular}{l ccc}
\rowcolor{headerrow}
\textbf{True} $\downarrow$ / \textbf{Pred.} $\rightarrow$ & Acc. & Esc. & Rej. \\
\hline
\multicolumn{4}{l}{\textit{1 Judge}} \\
Accept   & \cellcolor{diagcell}\textbf{7}  & \cellcolor{mildererr}1 & \cellcolor{criticalerr}2 \\
Escalate & \cellcolor{mildererr}0          & \cellcolor{diagcell}\textbf{9} & \cellcolor{mildererr}5 \\
Reject   & \cellcolor{criticalerr}0        & \cellcolor{mildererr}5 & \cellcolor{diagcell}\textbf{26} \\
\hline
\multicolumn{4}{l}{\textit{3 Judges}} \\
Accept   & \cellcolor{diagcell}\textbf{7}  & \cellcolor{mildererr}2 & \cellcolor{criticalerr}1 \\
Escalate & \cellcolor{mildererr}0          & \cellcolor{diagcell}\textbf{10} & \cellcolor{mildererr}4 \\
Reject   & \cellcolor{criticalerr}0        & \cellcolor{mildererr}6 & \cellcolor{diagcell}\textbf{25} \\
\hline
\multicolumn{4}{l}{\textit{5 Judges}} \\
Accept   & \cellcolor{diagcell}\textbf{5}  & \cellcolor{mildererr}3 & \cellcolor{criticalerr}2 \\
Escalate & \cellcolor{mildererr}0          & \cellcolor{diagcell}\textbf{10} & \cellcolor{mildererr}4 \\
Reject   & \cellcolor{criticalerr}0        & \cellcolor{mildererr}4 & \cellcolor{diagcell}\textbf{27} \\
\hline
\multicolumn{4}{l}{\textit{7 Judges}} \\
Accept   & \cellcolor{diagcell}\textbf{6}  & \cellcolor{mildererr}3 & \cellcolor{criticalerr}1 \\
Escalate & \cellcolor{mildererr}0          & \cellcolor{diagcell}\textbf{10} & \cellcolor{mildererr}4 \\
Reject   & \cellcolor{criticalerr}0        & \cellcolor{mildererr}4 & \cellcolor{diagcell}\textbf{27} \\
\end{tabular}
\end{ruledtabular}
\end{table}

\section{Conclusions}
\label{sec:conclusions}
We observe that this architecture for a verification layer presents an effective, operable, and environment-agnostic avenue for ensuring the safety, ethical alignment, and AI-specific and cyber-related security of proposed plans in a robot autonomy setting. In a significant majority of testing, errors concentrated on the escalate boundary, a weakness that does not hinder the most critical function of this system. While this work does not take advantage of domain-specific models or general-purpose large language models equipped with domain-specific skills, we encourage future research in this area. Another avenue for future work is applying a Conditional Value at Risk (CVaR) approach to provide a stronger mathematical grounding for this project and provide another decision input. Together, this would move the verification layer towards a robust, auditable, runtime structure that couples semantic safety considerations, high-level ethical and goal evaluation, and continuous capability-aware oversight within a single verification framework for future research in autonomy.

\begin{acknowledgments}
 This work was supported by the U.S. Department of Energy, Office of Science, Office of Workforce Development for Teachers and Scientists (WDTS) under the Science Undergraduate Labratory Interhsips Program (SULI).
\end{acknowledgments}

\section*{Code and Data Availability}
Code and datasets for the project can be found here: \url{https://github.com/rohanbhagra/verification_layers}
\section*{References}
\footnotesize
\setlength{\parskip}{2pt}
\setlength{\parindent}{0pt}

\hangindent=1.5em \hangafter=1
[1] A. Lynch \textit{et al.}, ``Agentic Misalignment: How LLMs Could Be Insider Threats,'' arXiv [cs.CR], 2025.

\hangindent=1.5em \hangafter=1
[2] T. Zhang \textit{et al.}, ``Using large language models for embodied planning introduces systematic safety risks,'' arXiv [cs.AI], 2026.

\hangindent=1.5em \hangafter=1
[3] R. Harang, ``Modeling Attacks on AI-Powered Apps with the AI Kill Chain Framework,'' NVIDIA Developer Blog, 2024. [Online]. 

\hangindent=1.5em \hangafter=1
[4] M. Ahn \textit{et al.}, ``AutoRT: Embodied Foundation Models for Large Scale Orchestration of Robotic Agents,'' arXiv [cs.RO], 2024.

\hangindent=1.5em \hangafter=1
[5] Y. Bai \textit{et al.}, ``Constitutional AI: Harmlessness from AI Feedback,'' arXiv:2212.08073 [cs.CL], 2022.

\hangindent=1.5em \hangafter=1
[6] X. Wang \textit{et al.}, ``Self-Consistency Improves Chain of Thought Reasoning in Language Models,'' arXiv [cs.CL], 2023.

\hangindent=1.5em \hangafter=1
[7] A. Panickssery, S. R. Bowman, and S. Feng, ``LLM Evaluators Recognize and Favor Their Own Generations,'' in \textit{Advances in Neural Information Processing Systems}, vol. 37, 2024, pp. 68772--68802.

\hangindent=1.5em \hangafter=1
[8] R. Zhao, W. Zhang, Y. K. Chia, W. Xu, D. Zhao, and L. Bing, ``Auto-Arena: Automating LLM Evaluations with Agent Peer Battles and Committee Discussions,'' arXiv:2405.20267 [cs.CL], 2024.

\hangindent=1.5em \hangafter=1
[9] F. Gr\"otschla, L. M\"uller, J. T\"onshoff, M. Galkin, and B. Perozzi, ``AgentsNet: Coordination and Collaborative Reasoning in Multi-Agent LLMs,'' arXiv:2507.08616 [cs.MA], 2025.

\hangindent=1.5em \hangafter=1
[10] A. D. Ames, S. Coogan, M. Egerstedt, G. Notomista, K. Sreenath, and P. Tabuada, ``Control Barrier Functions: Theory and Applications,'' arXiv:1903.11199 [eess.SY], 2019.

\hangindent=1.5em \hangafter=1
[11] S. Bansal, M. Chen, S. Herbert, and C. J. Tomlin, ``Hamilton-Jacobi Reachability: A Brief Overview and Recent Advances,'' arXiv:1709.07523 [cs.SY], 2017.

\hangindent=1.5em \hangafter=1
[12] A. A. Khan \textit{et al.}, ``Safety Aware Task Planning via Large Language Models in Robotics,'' arXiv [cs.RO], 2025.

\hangindent=1.5em \hangafter=1
[13] A. Kong \textit{et al.}, ``Better Zero-Shot Reasoning with Role-Play Prompting,'' arXiv:2308.07702 [cs.CL], 2024.

\hangindent=1.5em \hangafter=1
[14] T. Yuan \textit{et al.}, ``R-Judge: Benchmarking Safety Risk Awareness for LLM Agents,'' arXiv:2401.10019 [cs.CL], 2024.

\normalsize
\end{document}